\documentclass[sigconf]{acmart}
\AtBeginDocument{%
  }

\usepackage{multirow}
\usepackage{colortbl}
\usepackage{xcolor}
\definecolor{gray10}{gray}{0.9}

\usepackage{enumitem}
\usepackage{makecell}
\usepackage{pifont}
\definecolor{rowgray}{gray}{0.95}

\copyrightyear{2026}
\acmYear{2026}
\setcopyright{cc}
\setcctype{by}
\acmConference[MM '26]{Proceedings of the 34th ACM International Conference on Multimedia}{November 10--14, 2026}{Rio de Janeiro, Brazil}
\acmBooktitle{Proceedings of the 34th ACM International Conference on Multimedia (MM '26), November 10--14, 2026, Rio de Janeiro, Brazil}
\acmDOI{10.1145/3767308.3835644}
\acmISBN{979-8-4007-2213-4/2026/11}

\begin{document}

\title{SeqAlign3DVG: A Sequence-Aligned Benchmark and Voxel Reasoning Framework for 3D Visual Grounding}

\author{Yi Zhang}
\email{yi-eee.zhang@connect.polyu.hk}
\orcid{0009-0009-8242-1581}
\affiliation{%
  \institution{The Hong Kong Polytechnic University}
  \city{Hong Kong}
  \country{China}
}

\author{Yi Wang}
\correspondingauthor
\email{yi-eie.wang@polyu.edu.hk}
\orcid{0000-0001-8659-4724}
\affiliation{%
  \institution{The Hong Kong Polytechnic University}
  \city{Hong Kong}
  \country{China}
}

\author{Yueting Wu}
\email{lisala.wu@connect.polyu.hk}
\orcid{0009-0002-4739-7482}
\affiliation{%
  \institution{The Hong Kong Polytechnic University}
  \city{Hong Kong}
  \country{China}
}

\author{Kaiyue Yang}
\email{sheesage36@sjtu.edu.cn}
\orcid{0009-0000-6621-9206}
\affiliation{%
  \institution{Shanghai Jiao Tong University}
  \city{Shanghai}
  \country{China}
}

\author{Yuejiao Su}
\email{yuejiao.su@connect.polyu.hk}
\orcid{0009-0006-9118-9217}
\affiliation{%
  \institution{The Hong Kong Polytechnic University}
  \city{Hong Kong}
  \country{China}
}

\author{Lap-Pui Chau}
\correspondingauthor
\email{lap-pui.chau@polyu.edu.hk}
\orcid{0000-0003-4932-0593}
\affiliation{%
  \institution{The Hong Kong Polytechnic University}
  \city{Hong Kong}
  \country{China}
}

\renewcommand{\shortauthors}{Yi Zhang et al.}

\begin{abstract}
Image-based 3D visual grounding is critical for embodied agents, yet existing benchmarks suffer from loose text-observation alignment and neglect temporal ordering. We introduce SeqAlign3DVG, a novel benchmark dedicated to temporally ordered and strictly observation-aligned image-based 3D visual grounding. Unlike prior works using order-agnostic views or global point clouds, SeqAlign3DVG ensures all expressions are human-verified and strictly grounded in the provided RGB observations (single frames or ordered observation sequences). It comprises 9,622 single-view and 14,493 sequence samples featuring rich descriptions, complex relations, and multi-instance ambiguities. To tackle this benchmark, we propose a unified voxel-based pipeline featuring Relevance-Ordered Voxel Memory (ROVM) and Progressive Language-Voxel Fusion (PLVF). ROVM dynamically ranks and aggregates multi-view evidence via a conservative memory to mitigate noisy observations, while PLVF performs coarse-to-fine spatial-linguistic reasoning for precise disambiguation. Our approach achieves state-of-the-art performance under the depth-free protocol, significantly improving localization for targets defined by complex relations and appearance cues.
\end{abstract}



\begin{teaserfigure}
  \includegraphics[width=\textwidth]{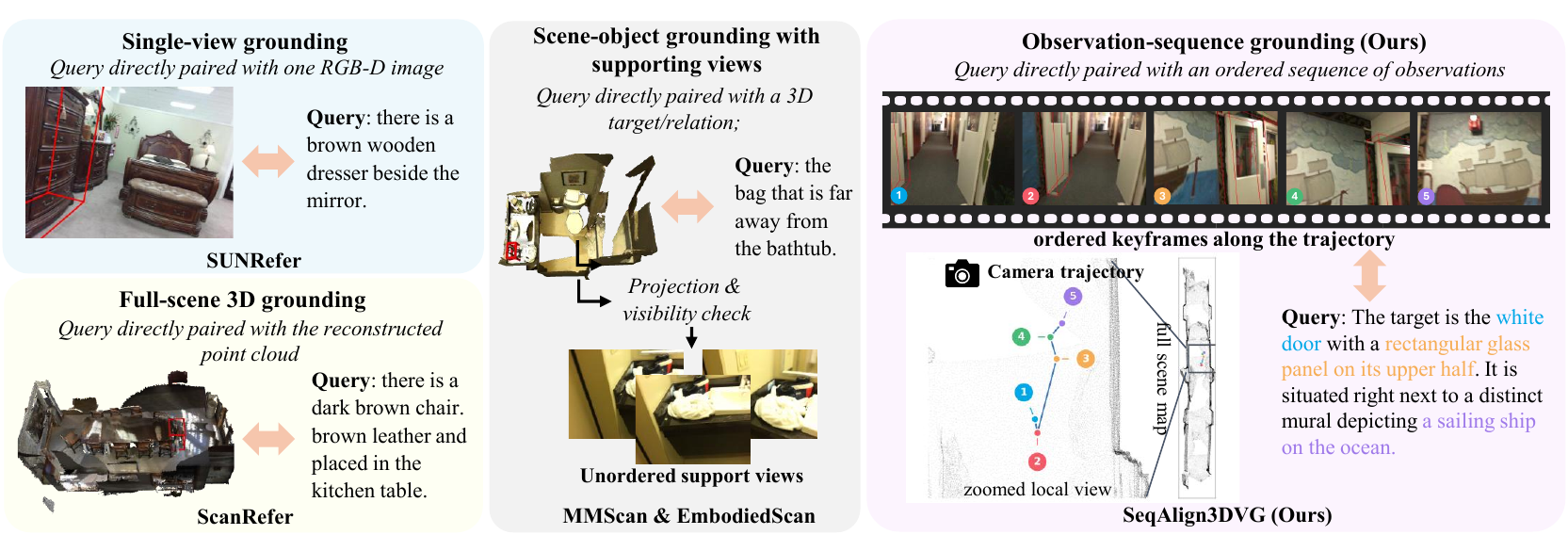}
  \caption{Benchmark taxonomy by what the query is directly paired with. Unlike prior works relying on global scenes or unordered views, our SeqAlign3DVG ensures strict alignment between queries and observation sequences or single views.}
  \Description{Existing grounding benchmarks pair queries with a single RGB-D image, a reconstructed full-scene point cloud, or a scene-level 3D target/relation whose supporting views are derived afterward via projection and visibility. In contrast, SeqAlign3DVG pairs each query with an ordered sequence of egocentric observations along the camera trajectory.}
  \label{fig:teaser}
\end{teaserfigure}


\maketitle

\section{Introduction}
\label{sec:intro}
Grounding free-form referring expressions to 3D objects is essential for embodied agents and AR systems to interact with the physical world, facilitating downstream tasks like navigation and manipulation. However, a critical gap exists between current 3D visual grounding benchmarks and realistic embodied perception. Most existing benchmarks \cite{chen2020scanrefer, Achlioptas2020ReferIt3DNL, kato2023arkitscenerefer, Zhang2023Multi3DReferGT} assume access to fully reconstructed global 3D scenes (e.g., complete point clouds or meshes). In practical indoor deployments, agents perceive environments through egocentric motion trajectories characterized by limited fields of view, partial observability, and viewpoint-dependent evidence. This discrepancy necessitates image-based 3D visual grounding: systems must infer a target's 3D location directly from sequential visual observations available during the perceptual process, rather than relying on a priori global reconstructions.

Existing image-based benchmarks, however, fall short of capturing these realistic dynamics. Early datasets are largely restricted to single-view grounding \cite{9578336} or outdoor driving scenarios \cite{Li2025NuGroundingAM, zhan2024mono3dvg}. While recent indoor embodied benchmarks, such as EmbodiedScan \cite{10655420} and MMScan \cite{10.5555/3737916.3739527}, have moved toward multi-view grounding, their paradigm for attaching language supervision introduces two fundamental limitations. In these datasets, expressions are typically associated with a scene-level target or a target-anchor relation, and the supporting views are subsequently selected through 3D-to-2D projection and visibility checks. Consequently, the language is not directly paired with the exact observations used for evaluation: described appearance cues or spatial evidence may be heavily occluded, ambiguous, or entirely absent in the provided frames, leading to a severe observability mismatch. Moreover, the retrieved views are treated as an unordered "bag of views" rather than a trajectory-preserving, temporally ordered observation sequence. This formulation neglects the temporal dynamics inherent in embodied perception, bypassing how agents progressively resolve visual ambiguities over time.

To address these critical issues, we introduce SeqAlign3DVG, an observation-aligned indoor benchmark for 3D visual grounding from both single views and observation sequences. SeqAlign3DVG supports two settings under a unified protocol: single-view grounding from an isolated RGB observation, and observation-sequence grounding from a temporally ordered RGB sequence. All expressions are human-verified against the exact evaluation observation(s), enforcing strict observation alignment and ensuring that the target can be uniquely identified from the provided visual evidence. Beyond alignment, SeqAlign3DVG is deliberately challenging: its expressions combine appearance cues and spatial relations, and a large fraction of the samples involve same-class distractors that require fine-grained disambiguation.

SeqAlign3DVG also highlights critical modeling challenges for image-based grounding. First, in realistic indoor trajectories, view quality is highly uneven: some frames capture clear target evidence, while others are heavily occluded or only observe supporting anchors. However, existing multi-view aggregation strategies typically treat inputs as order-agnostic sets and rely on uniform feature fusion. This inevitably blurs sharp geometric cues when integrating noisy or partial observations. Second, indoor queries often involve complex spatial relations and same-class distractors, requiring fine-grained token-level disambiguation. Standard 3D-language fusion mechanisms \cite{10655420, Li2025NuGroundingAM} typically apply dense global cross-attention between text and full 3D representations. This is computationally expensive and struggles to isolate subtle visual attributes in cluttered spaces. Therefore, a robust pipeline must intelligently filter uneven temporal evidence and perform focused, noise-reduced spatial-linguistic reasoning.

Motivated by this, we build a voxel-based grounding pipeline that supports both single-view and observation sequence inputs. Our pipeline lifts per-view 2D features into 3D voxel volumes using camera projection and regresses 3D bounding boxes directly. To effectively aggregate temporal sequences, we introduce a Relevance-Ordered Voxel Memory (ROVM) that ranks views by query relevance and updates a conservative memory to preserve high-quality evidence while suppressing noise. For fine-grained spatial-linguistic disambiguation, we propose Progressive Language-Voxel Fusion (PLVF), which establishes a coarse target hypothesis before applying sparse token-level cross-attention exclusively on the most promising voxels. We further stabilize training with an auxiliary occupancy supervision signal derived from ground-truth 3D boxes.


Empirically, our method achieves state-of-the-art performance under the depth-free protocol. On the observation-sequence split, it obtains 51.30\%/22.39\% Acc@0.25/0.50, exceeding BIP3D \cite{11093381} without ground-truth depth supervision but with detector initialization by 1.25/2.27 percentage points, while using neither target phrase-token supervision nor a pretrained 3D detector prior.

In summary, we make three contributions:
\begin{itemize}
    \item We introduce SeqAlign3DVG, an observation-aligned indoor benchmark for 3D visual grounding from single views and observation sequences, with human-verified expressions tied to the exact evaluation observation(s).
    \item We establish a unified voxel-based grounding pipeline and propose Relevance-Ordered Voxel Memory (ROVM) for query-aware sequence aggregation, alongside Progressive Language-Voxel Fusion (PLVF) for efficient spatial-linguistic disambiguation.
    \item We achieve state-of-the-art performance on SeqAlign3DVG under the depth-free protocol.
\end{itemize}

\section{Related Work}
\subsection{3D Visual Grounding Benchmarks}

\textbf{Reconstructed indoor 3D-scene benchmarks.} Most established 3D visual grounding (3DVG) benchmarks are built on reconstructed indoor scans, evaluating grounding with explicit 3D scene representations (e.g., point clouds or meshes). ScanRefer \cite{chen2020scanrefer} and ReferIt3D \cite{Achlioptas2020ReferIt3DNL} pioneered this field by providing free-form natural language and template-based relational utterances, respectively. Subsequent works like Multi3DRefer \cite{Zhang2023Multi3DReferGT}, ScanEnts3D \cite{10484480}, and ViGiL3D \cite{wang-etal-2025-vigil3d} further expanded the task to multi-target localization, phrase-to-object correspondences, and linguistically diverse diagnostics.

\begin{table*}[t]
\caption{Comparison with representative indoor 3D visual grounding benchmarks. Among the listed benchmarks, SeqAlign3DVG alone combines strict observation alignment, ordered sequences, and a unified setting for single-view and sequence grounding.}
\label{tab:benchmark_comparison}
\centering
\small
\setlength{\tabcolsep}{4pt}
\renewcommand{\arraystretch}{1.03}
\begin{tabular*}{\textwidth}{@{\extracolsep{\fill}}llccccc@{}}
\toprule
Dataset & Evaluation input & \makecell[c]{Exact evaluation-\\observation alignment$^\dagger$} & \makecell[c]{Ordered\\sequence} & \makecell[c]{Free-form\\language} & \makecell[c]{Single-view\\+ sequence} & Instances \\
\midrule
ScanRefer~\cite{chen2020scanrefer} & Global 3D scene & -- & \ding{55} & \ding{51} & \ding{55} & 11k \\
Nr3D~\cite{Achlioptas2020ReferIt3DNL} & Global 3D scene & -- & \ding{55} & \ding{51} & \ding{55} & 6k \\
SUNRefer~\cite{9578336} & Single RGB-D view & \ding{51} & \ding{55} & \ding{51} & \ding{55} & 7k \\
EmbodiedScan~\cite{10655420} & Multi-view set & \ding{55} & \ding{55} & \ding{55} & \ding{55} & 27k \\
MMScan~\cite{10.5555/3737916.3739527} & Multi-view set & \ding{55} & \ding{55} & \ding{51} & \ding{55} & 61k \\
\textbf{SeqAlign3DVG (ours)} & \textbf{Single view + sequence} & \ding{51} & \ding{51} & \ding{51} & \ding{51} & \textbf{$\sim$10k} \\
\bottomrule
\end{tabular*}

\parbox{\textwidth}{\footnotesize
\textbf{Instances} denote the number of unique target instances.
$^\dagger$ For image-based inputs, exact evaluation-observation alignment means that the released query is written and checked against the exact observation unit paired with that sample; ``--'' denotes that this view-level property is not applicable to a global 3D-scene input.
}
\end{table*}

\textbf{Image-based 3D grounding benchmarks.} Complementary to full-scan inputs, image-based benchmarks study grounding from posed camera views, where partial observability becomes a central challenge. Early datasets focused on single-view RGB-D grounding (SUNRefer \cite{9578336}) or outdoor driving scenarios (Mono3DRefer \cite{zhan2024mono3dvg}, NuGrounding \cite{Li2025NuGroundingAM}). Recently, embodied benchmarks like EmbodiedScan \cite{10655420} and MMScan \cite{10.5555/3737916.3739527} pushed toward ego-centric multi-view streams. 
However, their language is not necessarily written and verified against the exact evaluation observations, and their supporting frames are treated as order-agnostic sets. 
SeqAlign3DVG provides human-verified expressions strictly conditioned on the exact single-view or temporally ordered sequence inputs used for evaluation.

\subsection{3D Visual Grounding with Explicit 3D Geometry}
A broad line of work assumes access to explicit 3D geometry (e.g., point clouds with instance candidates). Many follow a grounding-by-matching paradigm, utilizing instance-level contextual modeling and relation reasoning \cite{9711198, He2021TransRefer3DEA, 10.5555/3600270.3601762}. Other works enhance robustness via multi-view projections \cite{9879926} or adopt detection-transformer-style formulations \cite{9878838, 10.1007/978-3-031-20059-5_24}. Beyond closed-set supervised 3DVG, recent work explores open-world zero-shot 3D grounding by leveraging large foundation models \cite{11094047, huang2025opengroundactivecognitionbasedreasoning}. Distinct from pre-scanned point clouds, another stream derives geometry from RGB-D inputs. Methods like SUNRefer \cite{9578336}, DenseGrounding \cite{zhengdensegrounding}, and DEGround \cite{zhang2025groundingdetectionenhancingcontextual} bridge 2D features with 3D spatial perception using depth maps. Since these approaches rely on explicit geometric reconstruction priors or depth inputs, they are not directly comparable to RGB-only grounding methods under depth-free settings.

\begin{figure*}[t]
    \centering
    \includegraphics[width=\textwidth]{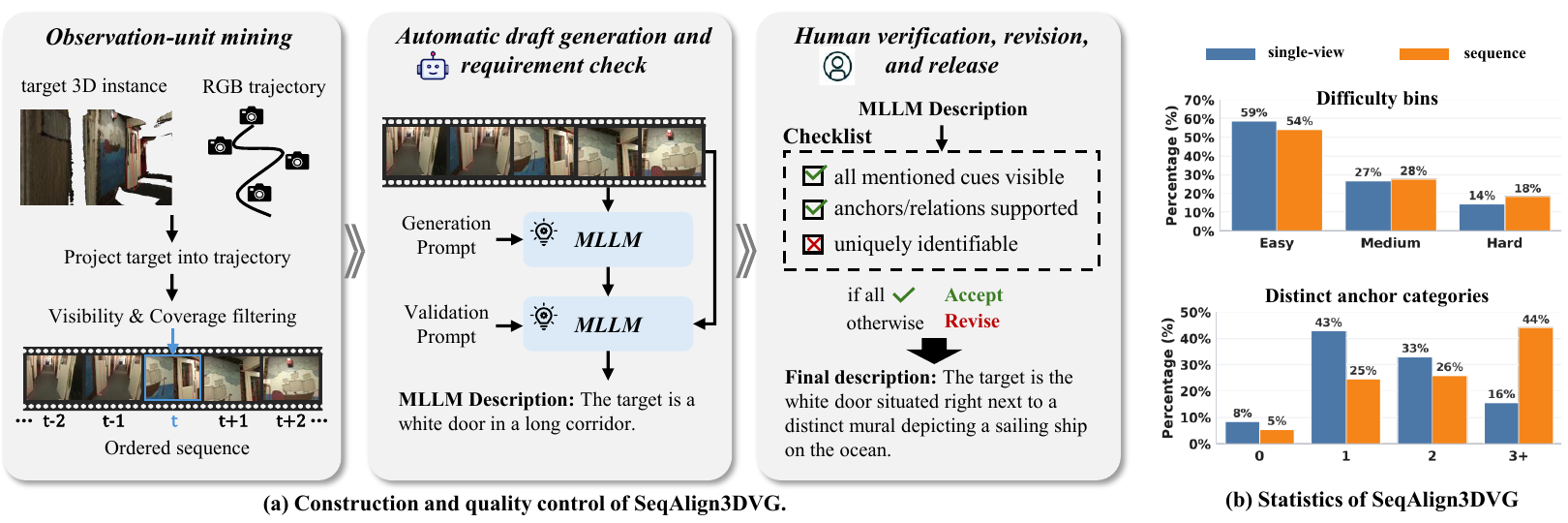}
    \caption{\textbf{(a) Construction pipeline.} Candidate views/sequences are paired with draft descriptions and human-verified against strict observability criteria. \textbf{(b) Dataset statistics,} illustrating the distribution of difficulty levels and anchor categories.}
    \Description{Two-part figure. Part (a) shows the SeqAlign3DVG construction pipeline from observation-unit mining to automatic description generation and human verification. Part (b) summarizes dataset statistics including difficulty bins and anchor-category distributions for the single-view and sequence splits.}
    \label{fig:benchmark_pipeline}
\end{figure*}

\subsection{RGB-only Image-based 3D Visual Grounding}
Compared with geometry-based 3DVG, RGB-only image-based 3D grounding avoids depth input at inference time and must infer 3D cues from visual observations. In monocular driving scenarios, Mono3DVG-TR \cite{zhan2024mono3dvg} proposes single-stage transformer-based monocular 3D grounding. Mono3DVG-EnSD \cite{li2025mono3dvgensdenhancedspatialawaredimensiondecoupled} further improves performance by decoupling dimension-specific textual features and dynamically masking high-certainty keywords to enforce spatial reasoning. For multi-view autonomous driving, NuGrounding \cite{Li2025NuGroundingAM} fine-tunes a Multimodal LLM to aggregate 3D geometric priors from BEV features with linguistic instructions for precise localization. 
However, these methods are predominantly tailored for outdoor driving, whereas BIP3D~\cite{11093381} supports indoor posed-image grounding with optional depth-map input through structured 2D features and explicit 3D position encoding. In our evaluation, its strongest depth-free variant uses target phrase-token supervision and detector initialization, and removing the latter causes a sharp performance drop; our depth-free framework instead uses shared voxel lifting, ROVM, and PLVF for strictly aligned sequences without either additional supervision or prior.

\section{The SeqAlign3DVG Benchmark}
\label{sec:benchmark}

SeqAlign3DVG is an indoor benchmark for image-based 3D visual grounding from both single RGB observations and temporally ordered observation sequences. Its central design principle is \emph{strict observation alignment}: each released expression is written and verified against the exact observation unit paired with that sample, whether a single view or an ordered sequence. This differs from prior benchmarks, where language is attached to scene-level target/anchor objects or relations, while the associated views are selected afterward according to visibility or benchmark protocol. As a result, exact text--observation alignment is not guaranteed at the sample level. SeqAlign3DVG also explicitly models temporal ordering and evaluates single-view and sequence grounding under a unified setting. Table~\ref{tab:benchmark_comparison} summarizes these distinctions.

\subsection{Construction and Quality Control}
\label{subsec:benchmark_construction}

SeqAlign3DVG is built on ScanNetV2 \cite{dai2017scannet} scenes with registered camera trajectories and aligned 3D instance annotations. We first mine candidate targets together with observation units from the original RGB streams. For the single-view split, an observation unit is one RGB frame in which the target is sufficiently visible and can in principle be uniquely identified from the selected view. For the sequence split, an observation unit is a temporally contiguous clip sampled along the original trajectory and preserved in its natural order. Crucially, language is attached to the selected observation unit itself, rather than inherited from a scene-level annotation or post-hoc support-view retrieval.

We use a scalable generate-then-verify pipeline. An automatic stage proposes candidate descriptions conditioned on the selected frame or sequence. Human annotators then verify each sample against the exact evaluation observation(s). A sample is accepted only if (1) the target is uniquely identifiable from the provided input, (2) every mentioned appearance cue, anchor object, and spatial relation is visually supported, and (3) the description does not rely on evidence unobservable anywhere in the provided input or on frames outside the evaluation input. Samples violating any of these criteria are rewritten. 
Figure~\ref{fig:benchmark_pipeline}a illustrates this pipeline and the strict verification criteria.

For a sequence sample, each mentioned target cue, anchor, and relation must be supported somewhere in the 15-frame input, but need not appear in every frame. Draft descriptions are generated with Qwen3-VL-32B \cite{bai2025qwen3vltechnicalreport} using structured prompts. Three annotators conduct independent, blind verification, with at least two annotators checking each sample; approximately 59\% of sequence drafts and 41\% of single-view drafts are rewritten. The train/validation split follows ScanNetV2's scene allocation, and the core sequence setting uses $T{=}15$ ordered keyframes.

\subsection{Dataset Characteristics and Diagnostic Subsets}
\label{subsec:benchmark_statistics}

The current core benchmark contains 9,622 single-view samples and 14,493 observation-sequence samples. These correspond to roughly 10k unique target instances. SeqAlign3DVG is intentionally designed to be diagnostic rather than merely large: many queries combine appearance cues with anchor objects and spatial relations, and a substantial fraction of samples involve same-class distractors that require fine-grained disambiguation.

To support detailed analysis, we organize the benchmark into diagnostic subsets along two axes, as summarized in Figure~\ref{fig:benchmark_pipeline}b. \textbf{Difficulty} is defined by the number of same-class distractors in the corresponding ScanNet scene, yielding \emph{Easy} $(0)$, \emph{Medium} $(1\text{--}2)$, and \emph{Hard} $(\geq 3)$ subsets. 
\textbf{Anchor usage} is characterized by the number of distinct anchor categories mentioned in each expression (0, 1, 2, or 3+). Descriptions with zero anchors rely primarily on intrinsic target cues. These subsets enable fine-grained evaluation of whether a model improves by better local appearance grounding, stronger relation reasoning, or more effective use of temporally accumulated evidence.

\section{Methodology}
\label{sec:method}

\subsection{Problem Definition and Overall Pipeline}
We address image-based 3D visual grounding under a unified setting that supports both single-view and observation-sequence inputs. Given a natural-language description \(L\), a sequence of RGB images \(\mathcal{I}=\{I_t\}_{t=1}^{T}\), camera intrinsics \(\mathcal{K}=\{\mathbf{K}_t\}_{t=1}^{T}\), and camera extrinsics \(\mathcal{T}=\{\mathbf{T}_t\}_{t=1}^{T}\), our goal is to localize the referred object with a 3D bounding box \(\mathbf{b}\in\mathbb{R}^{7}\). The sequence length is \(T=1\) for single-view grounding and \(T>1\) for observation-sequence grounding.

\begin{figure*}[t]
\centering
\includegraphics[width=0.999\linewidth]{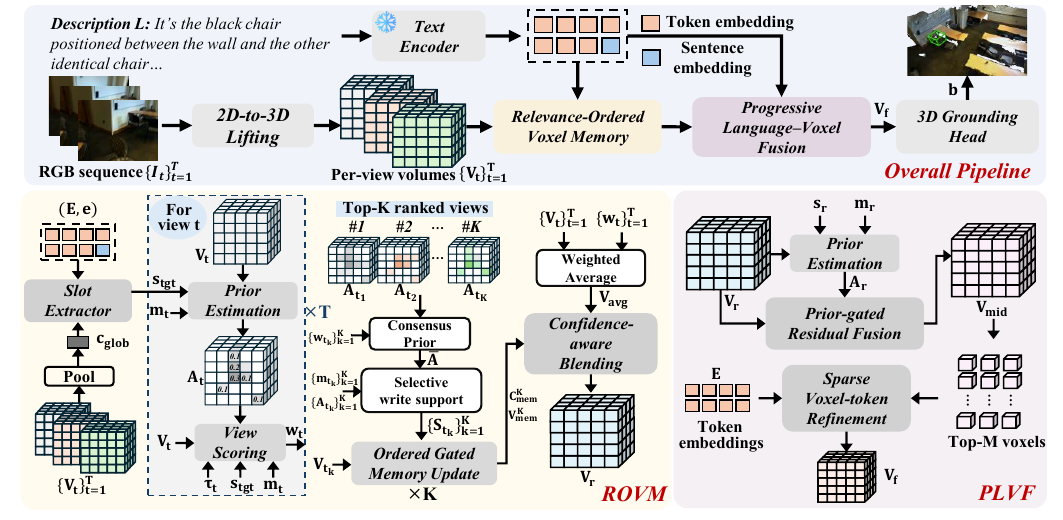}
\caption{Overall pipeline of our proposed method. It consists of four stages: (1) 2D-to-3D voxel lifting that maps RGB features into per-view volumes using camera projection; (2) Relevance-Ordered Voxel Memory (ROVM) for query-aware sequence aggregation; (3) Progressive Language-Voxel Fusion (PLVF) for fine-grained spatial-linguistic disambiguation; and (4) a 3D grounding head for final bounding box prediction.}
\Description{Overview of the proposed method. RGB observations are lifted to per-view voxel volumes, aggregated by the Relevance-Ordered Voxel Memory, refined by Progressive Language--Voxel Fusion, and decoded by a 3D grounding head to predict the target bounding box.}
\label{fig_framework}
\end{figure*}

As shown in Figure~\ref{fig_framework}, our pipeline bridges 1D language, 2D observations, and 3D geometric reasoning through a four-stage workflow:
(1) \textbf{2D-to-3D Voxel Feature Lifting}, which lifts each observation into a per-view 3D voxel volume using camera projection;
(2) \textbf{Relevance-Ordered Voxel Memory (ROVM)}, which is activated for observation sequences and consolidates the lifted per-view volumes into a query-relevant 3D representation;
(3) \textbf{Progressive Language--Voxel Fusion (PLVF)}, which performs coarse-to-fine language conditioning on the fused 3D volume; and
(4) a \textbf{3D Grounding Head} for final prediction.
For the single-view setting, Stage~(2) reduces to the identity mapping.

\subsection{2D-to-3D Voxel Feature Lifting}
We construct a unified 3D voxel grid \(\mathcal{V}\) of resolution \(X\times Y\times Z\), centered at the scene origin. A shared visual backbone extracts a 2D feature map \(\mathbf{F}_t\in\mathbb{R}^{C\times H\times W}\) from each image \(I_t\).

For each voxel center \(\mathbf{v}\in\mathcal{V}\), we obtain its projected 2D coordinate \(\mathbf{u}_t \in \mathbb{R}^2\) on the \(t\)-th image plane via standard camera matrix projection using intrinsics \(\mathbf{K}_t\) and extrinsics \(\mathbf{T}_t\). If \(\mathbf{u}_t\) lies inside the image bounds and the voxel is in front of the camera, we sample the 2D feature \(\mathbf{F}_t(\mathbf{u}_t)\) by bilinear interpolation.

Instead of collapsing the sequence at this stage, we retain a lifted voxel volume for each observation:
\begin{equation}
\mathbf{V}_t(\mathbf{v})=\mathbf{m}_t(\mathbf{v})\,\mathbf{F}_t(\mathbf{u}_t),
\label{eq:per_view_lift}
\end{equation}
where \(\mathbf{m}_t(\mathbf{v})\in\{0,1\}\) indicates whether voxel $\mathbf{v}$ is projectable into view $t$, i.e., its projection lies inside the image bounds and the voxel is in front of the camera.
This yields a set of per-view lifted volumes \(\{\mathbf{V}_t,\mathbf{m}_t\}_{t=1}^{T}\), where \(\mathbf{V}_t\in\mathbb{R}^{X\times Y\times Z\times C}\) and \(\mathbf{m}_t\in\{0,1\}^{X\times Y\times Z}\).

For single-view grounding (\(T=1\)), we directly set \(\mathbf{V}_r=\mathbf{V}_1\) and \(\mathbf{m}_r=\mathbf{m}_1\). For observation-sequence grounding (\(T>1\)), the set \(\{\mathbf{V}_t,\mathbf{m}_t\}_{t=1}^{T}\) is passed to the ROVM module for query-aware aggregation.

\subsection{Relevance-Ordered Voxel Memory}
\label{sec:rovm}
Given lifted per-view voxel volumes \(\{\mathbf{V}_t,\mathbf{m}_t\}_{t=1}^{T}\) and a language description \(L\), our goal is to aggregate the observation sequence in a query-aware manner. A key difficulty is that view usefulness is highly uneven: some views contain clear target evidence, whereas others are redundant, partially occluded, or only weakly related to the query. This issue is particularly severe for anchor-rich and relation-heavy expressions, where one view may clearly observe the target but not its supporting anchor, while another observes the anchor but only partial target evidence. Uniform fusion therefore tends to blur the cues needed for grounding. To address this, ROVM first estimates a target prior for each view, then ranks views by query relevance, and finally refines the sequence representation through a conservative memory written in relevance order.

\paragraph{Target-Slot-Guided Coarse Prior Estimation.}
To compare views meaningfully, we first need a coarse estimate of where the referred target may lie in each lifted volume. The description \(L\) is encoded into token embeddings \(\mathbf{E}\in\mathbb{R}^{N\times C}\) and a pooled sentence embedding \(\mathbf{e}\in\mathbb{R}^{C}\). Rather than matching every voxel against all tokens at this stage, we extract a compact target slot and use it as a stable target-oriented query for prior estimation. We first compute a pooled voxel context across the whole observation sequence:
\begin{equation}
\mathbf{c}_{\mathrm{glob}}
=
\frac{\sum_{t=1}^{T}\sum_{\mathbf{v}\in\mathcal{V}}
\mathbf{m}_t(\mathbf{v})\,\mathbf{V}_t(\mathbf{v})}
{\epsilon+\sum_{t=1}^{T}\sum_{\mathbf{v}\in\mathcal{V}}\mathbf{m}_t(\mathbf{v})}.
\end{equation}
We then obtain a context-aware target slot
\begin{equation}
\mathbf{s}_{\mathrm{tgt}}
=
g_{\mathrm{tgt}}\!\left(\mathbf{E},\mathbf{e},\phi_{\mathrm{ctx}}(\mathbf{c}_{\mathrm{glob}})\right),
\label{eq:target_slot}
\end{equation}
where \(g_{\mathrm{tgt}}\) is a lightweight slot extractor that uses a query conditioned on sentence-level text and pooled voxel context to attend over token embeddings and summarize them into a single target-focused text vector, and \(\phi_{\mathrm{ctx}}\) is a linear projection from voxel to text space.

For each view \(t\), we estimate a voxel-wise target prior by matching projected voxel features against \(\mathbf{s}_{\mathrm{tgt}}\):
\begin{align}
\mathbf{Z}_t(\mathbf{v})
&=
\beta\,
\mathrm{Sim}\!\left(
\Psi_{\mathrm{v2t}}(\mathbf{V}_t(\mathbf{v})),
\mathbf{s}_{\mathrm{tgt}}
\right), \nonumber\\
\mathbf{A}_t(\mathbf{v})
&=
\sigma(\mathbf{Z}_t(\mathbf{v}))\,\mathbf{m}_t(\mathbf{v}),
\label{eq:target_prior}
\end{align}
where \(\Psi_{\mathrm{v2t}}\) maps voxel features to the text space, \(\mathrm{Sim}(\cdot,\cdot)\) denotes similarity in the text space, \(\beta\) is a learnable logit scale, and \(\sigma(\cdot)\) is the sigmoid function.

\paragraph{Query-Aware View Scoring and Selection.}
To rank views, we summarize each lifted volume with a target-aware descriptor
\begin{equation}
\mathbf{d}_t
=
\frac{
\sum_{\mathbf{v}\in\mathcal{V}}
\bigl(\mathbf{A}_t(\mathbf{v})+\eta\mathbf{m}_t(\mathbf{v})\bigr)\mathbf{V}_t(\mathbf{v})
}{
\epsilon+\sum_{\mathbf{v}\in\mathcal{V}}
\bigl(\mathbf{A}_t(\mathbf{v})+\eta\mathbf{m}_t(\mathbf{v})\bigr)
},
\qquad
\mathbf{d}_t\in\mathbb{R}^{C},
\end{equation}
where \(\eta>0\) provides a small valid-region fallback when the prior is sparse. Thus, \(\mathbf{d}_t\) compactly summarizes the target-related content of view \(t\).

We additionally use two scalar reliability cues: a projection-coverage cue \(r_t=\mathrm{Mean}(\mathbf{m}_t)\), and a focus cue \(f_t=\mathrm{TopKMean}(\mathbf{A}_t\odot\mathbf{m}_t,K_f)\), where \(K_f\) is a small constant.
Here \(r_t\) measures how much of the voxel grid is projectable into the current view, whereas \(f_t\) reflects whether the target evidence is concentrated rather than diffuse. We also use a simple sequence-position cue \(\boldsymbol{\tau}_t=[p_t,1-p_t]\), where \(p_t=(t-1)/(T-1)\). These features are fed into a lightweight scorer:
\begin{equation}
\ell_t
=
g_{\mathrm{view}}([\mathbf{d}_t,\mathbf{s}_{\mathrm{tgt}},r_t,f_t,\boldsymbol{\tau}_t]),
\qquad
w_t
=
\left[\mathrm{MaskedSoftmax}(\{\ell_i\}_{i=1}^{T})\right]_t,
\label{eq:view_weight}
\end{equation}
where \(g_{\mathrm{view}}\) separately projects the target-aware view descriptor and the target slot into a shared hidden space, encodes the low-dimensional cues with a small MLP, and outputs a scalar relevance logit. The masked softmax is applied only over views with valid observations.

The resulting weights are also used to form a robust all-view baseline:
\begin{equation}
\mathbf{V}_{\mathrm{avg}}(\mathbf{v})
=
\frac{
\sum_{t=1}^{T} w_t\,\mathbf{m}_t(\mathbf{v})\,\mathbf{V}_t(\mathbf{v})
}{
\epsilon+\sum_{t=1}^{T} w_t\,\mathbf{m}_t(\mathbf{v})
}.
\label{eq:vavg}
\end{equation}

\paragraph{Relevance-Ordered Conservative Memory Refinement.}
View scoring identifies promising observations, but weighted averaging alone can blur sharp target regions by mixing uneven evidence. To address this, we refine the sequence using a conservative memory updated in descending order of view relevance, rather than the standard chronological order. This ensures the most confident views establish a strong initial target state, while subsequent, less confident views only selectively fill in missing details without overwriting reliable evidence or deviating far from the robust all-view baseline.

Let \(\mathcal{S} = (t_1, t_2, \dots, t_K)\) denote the ordered sequence of indices for the top-\(K\) valid views, sorted in descending order by their relevance scores \(\ell_t\). Within \(\mathcal{S}\), we renormalize the view weights as \(\tilde{w}_{t_k} = w_{t_k} / \sum_{i=1}^{K} w_{t_i}\). The resulting consensus prior is then computed as:
$$
\bar{\mathbf{A}}
=
\sum_{k=1}^{K}\tilde{w}_{t_k}\,\mathbf{A}_{t_k}.
$$

We initialize an empty memory volume, an accumulated observation mask, and a memory confidence map:
$$
\mathbf{V}_{\mathrm{mem}}^{(0)}=\mathbf{0}, \qquad
\mathbf{m}_{\mathrm{obs}}^{(0)}=\mathbf{0}, \qquad
\mathbf{C}_{\mathrm{mem}}^{(0)}=\mathbf{0}.
$$

The memory is then updated recurrently for \(k = 1, \dots, K\). For the \(k\)-th update step, directly using the raw prior \(\mathbf{A}_{t_k}\) as write support is risky, because broad low-confidence activations may overwrite the memory. We therefore derive a stricter write-support map:
$$
\mathbf{S}_{t_k}
=
\mathbf{m}_{t_k}
\odot
\mathbf{A}_{t_k}^{\gamma}
\odot
\bigl(0.5+0.5\bar{\mathbf{A}}\bigr),
\label{eq:write_support}
$$
where \(\gamma\ge 1\) is a sharpening exponent and \(\mathbf{A}_{t_k}^{\gamma}\) denotes element-wise exponentiation. This operation suppresses medium/low responses and retains compact high-confidence regions, while the consensus term favors regions repeatedly supported across selected views.

A lightweight gating network then predicts a voxel-wise update gate:
$$
\mathbf{U}_{t_k}=\Phi_{\mathrm{upd}}(\mathcal{G}_{t_k}),
\label{eq:update_gate}
$$
where \(\mathcal{G}_{t_k}\) concatenates the current memory \(\mathbf{V}_{\mathrm{mem}}^{(k-1)}\) and current-view features \(\mathbf{V}_{t_k}\), the target-support signals $\mathbf{S}_{t_k}$, the accumulated support state \((\mathbf{m}_{\mathrm{obs}}^{(k-1)},\mathbf{C}_{\mathrm{mem}}^{(k-1)})\), and simple update cues such as the update stage \(k/K\). The update is designed to favor reliable and previously unsupported target evidence.

To keep ROVM focused on selection and conservative correction rather than full language reasoning, we form a candidate write representation by adding only a weak target-aligned bias inside the supported region:
$$
\mathbf{H}_{t_k}
=
\mathbf{V}_{t_k}
+
\rho_{\mathrm{txt}}\,
\mathbf{S}_{t_k}\odot
\mathrm{Expand}(\mathbf{W}_{\mathrm{tv}}\mathbf{s}_{\mathrm{tgt}}),
$$
where \(\mathbf{W}_{\mathrm{tv}}\) projects the target slot to the voxel feature space, \(\rho_{\mathrm{txt}}\) is a learnable scalar gate, and \(\mathrm{Expand}(\cdot)\) broadcasts a vector to the voxel grid. Because this residual is restricted by \(\mathbf{S}_{t_k}\) and scaled by a small learnable gate, it only nudges voxels already supported as target evidence instead of rewriting the whole volume.

The memory states are then updated by:
\begin{align}
\mathbf{V}_{\mathrm{mem}}^{(k)}
&=
\mathbf{V}_{\mathrm{mem}}^{(k-1)}
+
\mathbf{U}_{t_k}\odot
\bigl(
\mathbf{H}_{t_k}-\mathbf{V}_{\mathrm{mem}}^{(k-1)}
\bigr), \\
\mathbf{m}_{\mathrm{obs}}^{(k)}
&=
\max(\mathbf{m}_{\mathrm{obs}}^{(k-1)},\mathbf{m}_{t_k}), \\
\mathbf{C}_{\mathrm{mem}}^{(k)}
&=
\max\!\left(
\mathbf{C}_{\mathrm{mem}}^{(k-1)},
\mathbf{S}_{t_k}\odot(0.5+0.5\mathbf{U}_{t_k})
\right).
\end{align}
Here \(\mathbf{V}_{\mathrm{mem}}^{(k)}\) stores the running refined memory, \(\mathbf{m}_{\mathrm{obs}}^{(k)}\) records which voxels have been covered by selected views up to step \(k\), and \(\mathbf{C}_{\mathrm{mem}}^{(k)}\) accumulates reliable target support for the final blending step.

After all \(K\) updates, we blend the refined memory with the robust all-view baseline:
$$
\mathbf{V}_r
=
\mathbf{V}_{\mathrm{avg}}
+
\sigma(\alpha_{\mathrm{blend}})\,
\mathbf{C}_{\mathrm{mem}}^{(K)}\odot
\bigl(
\mathbf{V}_{\mathrm{mem}}^{(K)}-\mathbf{V}_{\mathrm{avg}}
\bigr),
\label{eq:memory_blend}
$$
where \(\alpha_{\mathrm{blend}}\) is a learnable scalar. Consequently, confident target regions are sharpened by relevance-ordered memory refinement, while uncertain regions remain close to the robust all-view average.

We also keep the fused validity mask $\mathbf{m}_r$
which will be used by the subsequent PLVF module. ROVM thus outputs a query-relevant but only coarsely language-conditioned 3D volume, leaving finer token-level disambiguation to PLVF.

\subsection{Progressive Language--Voxel Fusion}
\label{sec:plvf}

ROVM effectively filters the observation sequence into a unified 3D volume with coarse spatial support. However, precise 3D visual grounding often hinges on fine-grained token-level cues, such as subtle visual attributes or complex target-anchor relations among nearby candidates. To bridge this gap, we introduce Progressive Language--Voxel Fusion (PLVF), a dedicated module that performs deep spatial-linguistic reasoning on the fused volume \((\mathbf{V}_r,\mathbf{m}_r)\). PLVF operates in a progressive two-stage manner: it first establishes a coarse target hypothesis using a condensed language slot, and subsequently revisits the full text tokens exclusively on the most promising voxels for fine-grained disambiguation.

\paragraph{Slot-Guided Coarse Target Hypothesis.}
Following the slot-based prior estimation formulation introduced in Sec.~\ref{sec:rovm}, but applied to the fused volume \((\mathbf{V}_r,\mathbf{m}_r)\), PLVF derives its own fused target slot \(\mathbf{s}_r\), prior logits $\mathbf{Z}_r$, and prior map \(\mathbf{A}_r\).

To explicitly emphasize candidate target regions, we use the fused prior to gate a lightweight residual:

\begin{equation}
\mathbf{V}_{\mathrm{mid}}
=
\mathbf{V}_{r}
+
\sigma(\gamma_1)\,
\Phi_{\mathrm{gate}}([\mathbf{V}_{r},\mathbf{A}_r,\mathbf{V}_{r}\odot\mathbf{A}_r]),
\end{equation}
where \(\Phi_{\mathrm{gate}}\) is a lightweight voxel fusion block and \(\gamma_1\) is a learnable scalar gate. This stage is computationally efficient, preserves the full 3D context, and supplies a coarse language-aware hypothesis for subsequent sparse reasoning.

\paragraph{Sparse Fine-Grained Token Refinement.}
The initial slot-guided stage relies on a compressed target slot \(\mathbf{s}_r\), which may miss fine-grained linguistic distinctions. We therefore bring back the full token sequence \(\mathbf{E}\), but only for a sparse set of candidate voxels. Specifically, after masking invalid voxels by \(\mathbf{m}_r\), we select the top-\(M\) voxel indices \(\mathcal{I}_{\mathrm{top}}\) according to the fused prior logits \(\mathbf{Z}_r\), and gather their features:
\begin{equation}
\mathbf{Q}_{\mathrm{sparse}}
=
\mathrm{Gather}(\mathbf{V}_{\mathrm{mid}},\mathcal{I}_{\mathrm{top}}),
\qquad
\mathbf{Q}_{\mathrm{sparse}}\in\mathbb{R}^{M\times C}.
\end{equation}
To preserve geometric structure, we project these voxel features to the text space and add 3D positional embeddings derived from normalized voxel coordinates:
\begin{equation}
\widetilde{\mathbf{Q}}_{\mathrm{sparse}}
=
\mathbf{W}_{q}\mathbf{Q}_{\mathrm{sparse}}
+
\mathbf{P}_{\mathrm{pos}}(\mathcal{I}_{\mathrm{top}}).
\end{equation}
We then perform multi-head cross-attention,
\begin{equation}
\mathbf{F}_{\mathrm{attn}}
=
\mathrm{MCA}(Q=\widetilde{\mathbf{Q}}_{\mathrm{sparse}},\,K=\mathbf{E},\,V=\mathbf{E}),
\end{equation}
so that only the most plausible target voxels re-examine the full text tokens. The attended features are mapped back to the voxel space and written to their original locations:
\begin{equation}
\mathbf{V}_f
=
\mathrm{Scatter}\!\left(
\mathbf{V}_{\mathrm{mid}},
\mathbf{Q}_{\mathrm{sparse}}+\sigma(\gamma_2)\mathbf{W}_{o}\mathbf{F}_{\mathrm{attn}},
\mathcal{I}_{\mathrm{top}}
\right),
\end{equation}
where \(\mathbf{W}_{o}\) maps the attended text features back to the voxel feature space and \(\gamma_2\) is a learnable scalar gate. This progressive design keeps expensive token-level reasoning off the full voxel grid and focuses it on a much cleaner candidate set produced by ROVM and the coarse prior-guided hypothesis.

\subsection{3D Grounding Head and Training Objectives}
The final fused volume \(\mathbf{V}_f\) is processed by a 3D feature pyramid network (FPN) to produce multi-scale representations. We employ a lightweight, anchor-free detection head shared across scales.

\textbf{Head Architecture.}
The head comprises two parallel branches: a score branch predicting a 3D confidence heatmap \(\hat{\mathbf{Y}}\in[0,1]^{X\times Y\times Z}\), and a regression branch predicting per-voxel box offsets representing the distances to the six faces of the target box.

\textbf{Loss Functions.}
The network is trained end-to-end with a composite objective \(\mathcal{L} = \mathcal{L}_{\mathrm{score}} + \mathcal{L}_{\mathrm{bbox}} + \lambda_{\mathrm{prior}}\mathcal{L}_{\mathrm{prior}} + \lambda_{\mathrm{view}}\mathcal{L}_{\mathrm{view}}\). Specifically, the score loss (\(\mathcal{L}_{\mathrm{score}}\)) applies Gaussian Focal Loss to supervise the predicted heatmap \(\hat{\mathbf{Y}}\), where the ground-truth heatmap is obtained by splatting a Gaussian kernel at the target center. The bounding box loss (\(\mathcal{L}_{\mathrm{bbox}}\)) applies an IoU loss to predicted boxes at positive voxel locations. To encourage meaningful coarse localization, the prior supervision (\(\mathcal{L}_{\mathrm{prior}}\)) minimizes the binary cross-entropy between the fused prior map \(\mathbf{A}_r\) and an auxiliary binary occupancy grid \(\mathbf{O}_{\mathrm{gt}}\in\{0,1\}^{X\times Y\times Z}\). Notably, \(\mathbf{O}_{\mathrm{gt}}\) is directly derived from the ground-truth 3D bounding box without requiring any additional signals. Finally, for observation-sequence grounding, the view-relevance supervision (\(\mathcal{L}_{\mathrm{view}}\)) applies a weighted soft binary cross-entropy to the view logit \(\ell_t\) in Eq.~\eqref{eq:view_weight}. It is supervised by a soft target label \(y_t \in [0,1]\), defined as the fraction of the ground-truth target volume covered by view \(t\). We assign larger weights to views with larger target coverage, and this loss is applied exclusively in the observation-sequence setting.

\section{Experiments}
\label{sec:experiments}

\begin{table*}[t]
\centering
\caption{\textbf{Observation-sequence Performance.} Protocol-aware comparison on the SeqAlign3DVG observation-sequence split.}
\label{tab:sequence_results}

\normalsize
\setlength{\tabcolsep}{5.0pt}
\renewcommand{\arraystretch}{1.03}

\begin{tabular}{@{}lcccccc@{}}
\toprule
\textbf{Method}
& \makecell[c]{\textbf{GT Depth}\\\textbf{Input}}
& \makecell[c]{\textbf{GT Depth}\\\textbf{Sup.}}
& \makecell[c]{\textbf{Phrase}\\\textbf{Sup.}}
& \makecell[c]{\textbf{3D Det.}\\\textbf{Prior}}
& \textbf{Acc@0.25}
& \textbf{Acc@0.50} \\
\midrule

Detector + Top-Score
& \ding{55} & \ding{55} & \ding{55} & \ding{51}
& 21.26 & 14.30 \\

Detector + CLIP
& \ding{55} & \ding{55} & \ding{55} & \ding{51}
& 31.86 & 18.79 \\

Detector + MiniLM-L6-v2
& \ding{55} & \ding{55} & \ding{55} & \ding{51}
& 30.01 & 17.49 \\

\midrule

Grounding DINO + Back-project
& \ding{51} & \ding{55} & \ding{55} & \ding{55}
& 26.61 & 5.62 \\

VLM-Grounder$^*$
& \ding{51} & \ding{55} & \ding{55} & \ding{55}
& 47.28 & 21.18 \\

BIP3D (w/o Depth Sup. \& Det. Init.)
& \ding{55} & \ding{55} & \ding{51} & \ding{55}
& 5.69 & 0.24 \\

BIP3D (w/o Depth Sup.)
& \ding{55} & \ding{55} & \ding{51} & \ding{51}
& 50.05 & 20.12 \\

BIP3D (w/ Depth Sup.)
& \ding{55} & \ding{51} & \ding{51} & \ding{51}
& 59.60 & 24.55 \\

\midrule

\rowcolor{gray10}
\textbf{Ours}
& \ding{55} & \ding{55} & \ding{55} & \ding{55}
& \textbf{51.30} & \textbf{22.39} \\

\bottomrule
\end{tabular}

\vspace{0.4mm}
\parbox{0.96\textwidth}{\scriptsize
\textbf{GT Depth Input}: ground-truth depth maps used at inference.
\textbf{GT Depth Sup.}: ground-truth depth used as training supervision without depth-map input at inference.
\textbf{Phrase Sup.}: target phrase-token supervision.
\textbf{3D Det. Prior}: proposals from a pretrained 3D detector or initialization from a pretrained 3D detection checkpoint.
$^*$ Our VLM-Grounder adaptation uses GPT-5.4-mini.
Bold denotes the best result under the depth-free protocol, i.e., without ground-truth depth input or depth supervision.
}
\end{table*}

\begin{table}[t]
\centering
\caption{\textbf{Single-view Performance.} Overall comparison on the SeqAlign3DVG single-view split.}
\label{tab:single_view_results}

\normalsize
\setlength{\tabcolsep}{3.0pt}
\renewcommand{\arraystretch}{1.03}

\begin{tabular}{@{}lcc@{}}
\toprule
\textbf{Method}
& \textbf{Acc@0.25}
& \textbf{Acc@0.50} \\
\midrule

Detector + Top-Score
& 22.97 & 9.87 \\

Detector + CLIP
& 29.44 & 11.28 \\

Detector + MiniLM-L6-v2
& 27.73 & 10.69 \\

\midrule

Grounding DINO + Back-project$^{\dagger}$
& 38.25 & 8.52 \\

\midrule

\rowcolor{gray10}
\textbf{Ours}
& \textbf{44.54}
& \textbf{16.86} \\

\bottomrule
\end{tabular}

\vspace{0.3mm}
\parbox{0.98\columnwidth}{\scriptsize
$^{\dagger}$ Uses ground-truth depth maps at inference.
}
\end{table}

\subsection{Experimental Settings}

\textbf{Datasets and Metrics.}
We evaluate our method on the proposed SeqAlign3DVG benchmark, reporting results on both the single-view and observation-sequence splits. Following standard protocols, we adopt Acc@0.25 and Acc@0.50 as our primary metrics, which measure the percentage of predicted bounding boxes having an Intersection-over-Union (IoU) with the ground truth strictly greater than 0.25 and 0.50, respectively.

\textbf{Implementation Details.}
We implement our pipeline using the MMDetection3D framework. The 3D voxel grid covers a spatial range of $6.4\,\mathrm{m} \times 6.4\,\mathrm{m} \times 2.56\,\mathrm{m}$ with a voxel resolution of $0.16\,\mathrm{m}$. The model is trained for 12 epochs using the AdamW optimizer with an initial learning rate of $5 \times 10^{-5}$, which is decayed by a factor of 10 at epochs 8 and 11. The loss weights for the auxiliary objectives are set to $\lambda_{prior}=0.5$ and $\lambda_{view}=0.05$. We set $M=512$ and $K=5$. All experiments are conducted on 3 NVIDIA RTX 6000 Ada GPUs.

\subsection{Baselines}
For the observation-sequence split, SeqAlign3DVG evaluates strictly observation-aligned grounding from ordered posed-RGB inputs. Among the evaluated methods, BIP3D provides the closest end-to-end posed-image comparison. However, the methods differ in depth input or supervision, target phrase-token supervision, and 3D detector priors, which are made explicit in Table~\ref{tab:sequence_results}.

\textbf{Detector-based re-ranking.}\ 
We use MVSDet~\cite{10.5555/3737916.3742138}, pretrained on the ScanNetV2 training split, to generate candidate 3D proposals and rank them by detector objectness (Top-Score), CLIP image--text similarity~\cite{Radford2021LearningTV}, or MiniLM-L6-v2 class--text similarity. This detector-agnostic pipeline could in principle be instantiated with other image-based 3D detectors, such as 3DGeoDet~\cite{11045444} and GVSynergy-Det~\cite{zhang2025gvsynergydetsynergisticgaussianvoxelrepresentations}.

\textbf{Other grounding baselines.}\ Grounding DINO + Back-project~\cite{10.1007/978-3-031-72970-6_3} and VLM-Grounder~\cite{pmlr-v270-xu25c} use ground-truth depth maps at inference to recover 3D boxes; our VLM-Grounder adaptation uses GPT-5.4-mini. In our experiments, all three BIP3D~\cite{11093381} variants receive posed RGB images without depth-map input, while ground-truth depth supervision and detector initialization are independently varied.

\subsection{Main Results}

Tables~\ref{tab:sequence_results} and~\ref{tab:single_view_results} present results for the observation-sequence and single-view settings, respectively. Our method attains the best performance under the depth-free protocol in both settings; methods using ground-truth depth input or depth supervision are included with explicit protocol labels for transparency.

\textbf{Observation-sequence Grounding.}
As shown in Table~\ref{tab:sequence_results}, our method achieves state-of-the-art performance under the depth-free protocol, reaching 51.30\%/22.39\% Acc@0.25/0.50. It outperforms BIP3D without depth supervision but with detector initialization (50.05\%/20.12\%) by 1.25/2.27 percentage points, while using neither target phrase-token supervision nor a pretrained 3D detector prior. Without detector initialization, BIP3D drops to 5.69\%/0.24\%, indicating a strong reliance on this prior on SeqAlign3DVG. VLM-Grounder obtains 47.28\%/21.18\% using ground-truth depth maps at inference, whereas BIP3D with depth supervision reaches 59.60\%/24.55\% without depth-map input; both settings lie outside the depth-free protocol. Grounding DINO + Back-project likewise uses ground-truth depth input but reaches only 26.61\%/5.62\%, indicating that independent frame-wise grounding followed by direct back-projection remains insufficient for accurate sequence-level 3D localization.

\textbf{Single-view Grounding.}
Table~\ref{tab:single_view_results} reports the overall single-view comparison. Without ground-truth depth input or depth supervision, our method achieves 44.54\% Acc@0.25 and 16.86\% Acc@0.50. Grounding DINO + Back-project uses ground-truth depth maps at inference but reaches 38.25\%/8.52\%, indicating that direct 2D grounding and depth-based back-projection remain insufficient for precise 3D localization. 

\subsection{Ablation Study}

\begin{table}[t]
\centering
\caption{\textbf{Component Ablation.} Incremental contributions of PLVF and ROVM.}
\label{tab:component_ablation}
\small
\setlength{\tabcolsep}{3.2pt}
\renewcommand{\arraystretch}{1.02}

\resizebox{\columnwidth}{!}{%
\begin{tabular}{@{}lcccc@{}}
\toprule
\textbf{Model}
& \multicolumn{2}{c}{\textbf{Single-view}}
& \multicolumn{2}{c}{\textbf{Sequence}} \\
\cmidrule(lr){2-3}
\cmidrule(l){4-5}
& \textbf{Acc@0.25}
& \textbf{Acc@0.50}
& \textbf{Acc@0.25}
& \textbf{Acc@0.50} \\
\midrule

Baseline
& 38.25 & 13.40
& 41.46 & 17.73 \\

+ PLVF
& \textbf{44.54} & \textbf{16.86}
& 46.09 & 20.40 \\

+ ROVM (Full)
& -- & --
& \textbf{51.30} & \textbf{22.39} \\

\bottomrule
\end{tabular}%
}
\end{table}

To validate that the architecture is supported by controlled design choices rather than a generic stack of layers, we conduct incremental component ablations in Table~\ref{tab:component_ablation}. The \textit{Baseline} removes both ROVM and PLVF.

\textbf{Effectiveness of PLVF.}
Integrating PLVF improves Acc@0.25 by 6.29 points on single-view and 4.63 points on sequence grounding, confirming the value of progressive coarse-to-fine voxel--token reasoning. Compared with dense full-grid cross-attention, PLVF reduces single-view computation by 35.7\% (158.69 to 101.99 GFLOPs), while lowering latency from 63 to 59 ms and peak memory from 3026 to 2810 MB.

\textbf{Effectiveness of ROVM.}
Adding ROVM raises sequence performance from 46.09\%/20.40\% to 51.30\%/22.39\%. 
In an additional write-order comparison, relevance-ordered writing outperforms the strongest alternative, reverse writing (50.34\%/21.87\%), by 0.96/0.52 points, supporting the deliberate ordering and conservative memory design.

\section{Conclusion}
We introduce SeqAlign3DVG, a strictly observation-aligned indoor benchmark for 3D grounding from single views and temporally ordered observation sequences. We further propose a unified voxel-based framework with ROVM for query-aware sequence aggregation and PLVF for efficient coarse-to-fine language--voxel reasoning. Extensive experiments demonstrate state-of-the-art performance under the depth-free protocol.

\begin{acks}
The research work described in this paper was conducted in the JC STEM Lab of Machine Learning and Computer Vision funded by The Hong Kong Jockey Club Charities Trust. This research was partially supported by a grant from the Research Grant Council, Hong Kong Special Administrative Region, China, under Project PolyU 15215824 and received partially support from the Global STEM Professorship Scheme from the Hong Kong Special Administrative Region.
\end{acks}

\bibliographystyle{ACM-Reference-Format}
\balance
\bibliography{sample-base}


\end{document}